\relax
\documentclass[letterpaper]{article} 
\usepackage{aaai19}  
\usepackage{times}  
\usepackage{helvet}  
\usepackage{courier}  
\usepackage{url}  
\usepackage{graphicx}  
\usepackage{amsmath}
\usepackage{array}
\usepackage{multirow}
\usepackage{algorithm}
\usepackage{algorithmic}
\usepackage{bm}
\usepackage{amssymb}

\newcommand{\argmax}{\mathop{\arg\!\max}}

\def\comment#1{{}}
\def\eg{{\em e.g.}}
\def\ie{{\em i.e.}}
\def\etal{{\em et al.}}

\begin{document}
%
\title{Learning Non-Uniform Hypergraph for Multi-Object Tracking}
\author{Longyin Wen$^1$\thanks{Equal contribution.}, Dawei Du$^2$\footnotemark[1], Shengkun Li$^2$, Xiao Bian$^3$, Siwei Lyu$^2$\\
$^1$JD Finance, Mountain View, CA, USA.\\
$^2$University at Albany, State University of New York, NY, USA.\\
$^3$GE Global Research, NY, USA\\
}
\maketitle
\begin{abstract}
The majority of Multi-Object Tracking (MOT) algorithms based on the tracking-by-detection scheme do not use higher order dependencies among objects or tracklets, which makes them less effective in handling complex scenarios. In this work, we present a new near-online MOT algorithm based on non-uniform hypergraph, which can model different degrees of dependencies among tracklets in a unified objective. The nodes in the hypergraph correspond to the tracklets and the hyperedges with different degrees encode various kinds of dependencies among them. Specifically, instead of setting the weights of hyperedges with different degrees empirically, they are learned automatically using the structural support vector machine algorithm (SSVM). Several experiments are carried out on various challenging datasets (\ie, PETS09, ParkingLot sequence, SubwayFace, and MOT16 benchmark), to demonstrate that our method achieves favorable performance against the state-of-the-art MOT methods.
\end{abstract}

\section{Introduction}
Multi-object tracking (MOT) is an important problem in computer vision with many applications, such as surveillance, behavior analysis, and sport video analysis. Although the performance of MOT has been significantly improved in recent years \cite{DBLP:conf/iccv/Choi15,DBLP:conf/iccv/KimLCR15,DBLP:journals/pami/WenLLL016,DBLP:conf/cvpr/SyT17}, it is still a challenging problem due to factors such as missed detections, false detections, and identification switches.

\begin{figure}[t]
\centering
\includegraphics[width=0.95\linewidth]{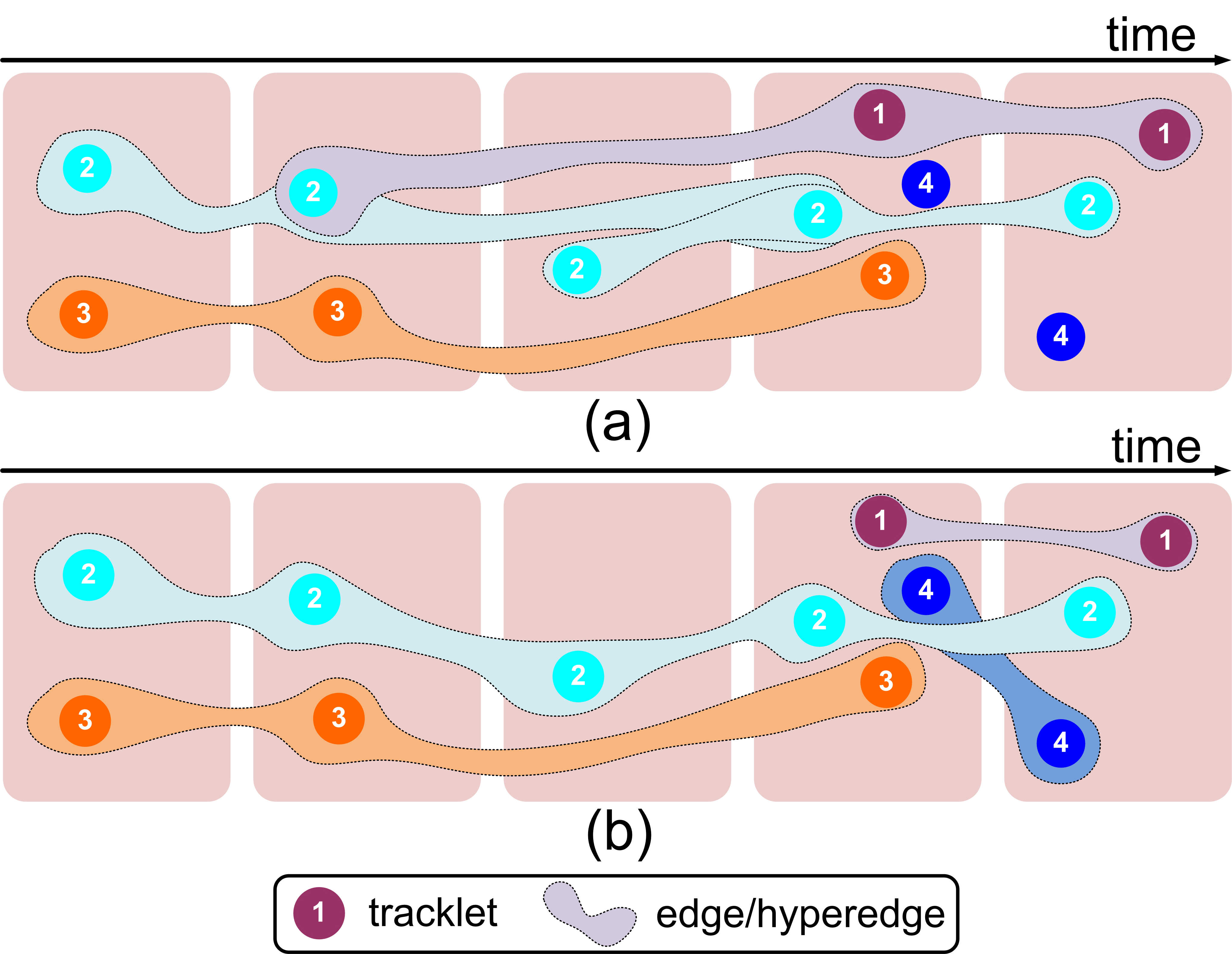}
\vspace{-3mm}
\caption{\small{(a) Two previous methods using $3$-uniform hypergraph H\textsuperscript{2}T \cite{DBLP:conf/cvpr/Longyin14} and FH\textsuperscript{2}T \cite{DBLP:journals/pami/WenLLL016}, often fails to describe the dependencies among tracklets, when occlusion or missed detection happen. (b) The proposed method uses the non-uniform hypergraph to encode different degrees of dependencies among tracklets effectively.}}
\label{fig:mot}
\end{figure}

An automatic MOT system usually employs a pre-trained object detector to locate candidate object regions in each frame, then match the detections across frames to form target trajectories. Most existing methods only consider the pairwise dependencies of detections (\eg, \cite{DBLP:conf/iccv/RezatofighiMZSD15,DBLP:conf/cvpr/DehghanAS15,DBLP:journals/pami/MilanSR16,DBLP:conf/eccv/McLaughlinRM16}), and do not take full advantage of the high-order dependencies among multiple targets across frames. This strategy is less effective when nearby objects with similar appearance or motion patterns occlude each other in the video. Several recent methods \cite{DBLP:conf/iccv/KimLCR15,DBLP:conf/cvpr/Collins12,DBLP:conf/cvpr/ShiLHYX14,DBLP:conf/iccv/KimLCR15,DBLP:conf/cvpr/Longyin14,DBLP:journals/pami/WenLLL016} attempt to use the high-order information to improve the tracking performance, such as dense structure search on hypergraph \cite{DBLP:conf/cvpr/Longyin14,DBLP:journals/pami/WenLLL016}, tensor power iterations \cite{DBLP:conf/cvpr/ShiLHYX14}, high-order motion constraints \cite{DBLP:conf/cvpr/Collins12,DBLP:conf/cvpr/ButtC13}, and multiple hypothesis tracking \cite{DBLP:conf/iccv/KimLCR15}. However, the aforementioned methods merely exploit fixed degrees of dependencies among objects, which limits the flexibility of the hypergraph model\footnote{A hypergraph is a generalization of a conventional graph where an edge can join more than two nodes.} in complex environments, and calls for adaptive dependency patterns. As shown in Figure \ref{fig:mot}, $3$-uniform hypergraph is unable to describe the dependencies between two tracklets of target $1$ and $4$ correctly. On the contrary, non-uniform hypergraph better adapts to different degrees of dependencies among tracklets, and achieves more reliable performance.

In this paper, we describe a new {\em non-uniform} hypergraph learning based tracker (NT), which has much stronger descriptive power to accommodate different tracking scenarios than the conventional graph \cite{DBLP:conf/cvpr/DehghanAS15} or uniform hypergraph \cite{DBLP:conf/cvpr/Longyin14,DBLP:journals/pami/WenLLL016}. The nodes in the hypergraph correspond to the tracklets\footnote{The terminology ``tracklet'' indicates a fragment of target trajectory. Notably, the input detection responses in each frame can be treated as tracklets of length one.}, and the hyperedges with different degrees encode similarities among tracklets to assemble various kinds of appearance and motion patterns. The tracking problem is formulated as searching dense structures on the non-uniform hypergraph. Different from previous methods \cite{DBLP:conf/cvpr/Longyin14,DBLP:journals/pami/WenLLL016}, we do not fix the degree of the hypergraph model, but mix hyperedges of different degrees and learn their relative weights automatically from the data using the structural support vector machine (SSVM) method \cite{DBLP:journals/ml/JoachimsFY09}. We propose an efficient approximation algorithm to exploit the dense structures to form long object trajectories to complete the tracking task. In addition, to achieve both accuracy and efficiency, we use a near-online strategy for MOT, \ie, we perform the dense structure searching on the non-uniform hypergraph to generate short tracklets in a temporal window, and then associate those short tracklets to the tracked targets to get the final trajectories of targets at the current time stamp. This process is carried out repeatedly to complete the tracking task in a video.

The main contributions are summarized as follows. (1) We propose a non-uniform hypergraph learning based near-online MOT method, which assembles the hyperedges with different degrees to encode various types of dependencies among objects. (2) The weights of hyperedges with different degrees in the non-uniform hypergraph are learned from data using the SSVM algorithm. (3) We propose an efficient approximation algorithm to complete the dense structure searching problem on the non-uniform hypergraph.

\section{Related Work}
MOT methods can be roughly classified into three categories, 1) online strategy, 2) off-line processing strategy, and 3) near-online strategy. If there occurs an error in tracking, it is hard for online strategy (\eg, \cite{DBLP:conf/cvpr/YangLWYL14,DBLP:conf/iccv/XiangAS15,DBLP:conf/cvpr/JHYCMK16}) to recover from due to imprecise appearance or motion measurements. Thus, many algorithms focus on off-line strategy (\eg, \cite{DBLP:journals/pami/BerclazFTF11,DBLP:conf/cvpr/SyT17,DBLP:journals/pami/MilanSR16}). To make the association step efficient, \cite{DBLP:journals/pami/BerclazFTF11} formulate the association as a constrained flow optimization problem, solved by the k-shortest paths algorithm. \cite{DBLP:conf/cvpr/SyT17} present a graph-based formulation that links and clusters person hypotheses over time by solving an instance of a minimum cost lifted multicut problem. In addition, Milan~\etal~\cite{DBLP:journals/pami/MilanSR16} pose MOT as minimization of a unified discrete-continuous energy function using the L-BFGS and QPBO algorithms. However, as only association between pairs of detections in local temporal domain are considered, the aforementioned methods do not perform well when multiple similar objects appear in proximity with clutter backgrounds.

To alleviate this problem, \cite{DBLP:conf/cvpr/DehghanAS15} use a graph to integrate all the relations among objects in a batch of frames and formulate the MOT problem as a Generalized Maximum Multi Clique problem on the graph. \cite{DBLP:conf/cvpr/Longyin14} exploit the motion information to help tracking and formulate MOT as the dense structure searching on a uniform hypergraph, in which the nodes correspond to tracklets and the edges encode the high-order dependencies among tracklets. To further improve the efficiency, an approximate RANSAC-style approach is proposed in \cite{DBLP:journals/pami/WenLLL016} to complete the dense structure searching.

Besides, \cite{DBLP:conf/iccv/Choi15} designs a near-online strategy, which inherits the advantages of both online and offline approaches. The tracking problem is formulated as a data-association between targets and detections in a temporal window, that is performed repeatedly at every frame. In this way, the algorithm is able to fix any association error made in the past when more detections are provided. \cite{DBLP:conf/bmvc/WangF15} present an end-to-end framework to learn parameters of min-cost flow for MOT problem using a tracking-specific loss function in the SSVM framework. Nevertheless our approach uses the non-uniform hypergraph to describe the high-order dependencies among tracklets, and uses SSVM framework to learn the weights of the hyperedges with different degrees.

\section{Non-uniform Hypergraph}
{\noindent {\bf Definition.}}
A hypergraph is a generalization of a conventional graph, where an edge can join more than two nodes. We use ${\cal G}({\it V},{\cal E}, {\cal A})$ to denote a (weighted) hypergraph, where ${\it V}=\{v_1,\cdots,v_{n}\}$ is the node set, $v_i$ is the $i$-th node and $n$ is the total number of nodes, ${\cal E}$ is the set of hyperedges, and ${\cal A}$ is the affinity set corresponding to the edges/hyperedges. Specifically, we define ${\cal E}={\it E}_{1}\cup\cdots\cup{\it E}_{D}$, where ${\it E}_1=\{(v_1),\cdots,(v_n) \}$ is the set of self-loops, ${\it E}_2\subseteq{\it V}\times{\it V}$ is the set of conventional graph edges, ${\it E}_{d}\subseteq{\it V}^{d}$ is the set of hyperedges with degree $d$, $d=3,\cdots,D$, and $D$ is the maximal degree of hyperedges. If all hyperedges in ${\cal G}$ have the same cardinality $d$, ${\cal G}$ is a $d$-uniform hypergraph (\ie, ${\it E}_{d'}=\emptyset$ for $d'\ne{d}$); otherwise, ${\cal G}$ is a non-uniform hypergraph. For node $v$, we denote its neighborhood as ${\cal N}(v)$, which is the set of nodes connected to $v$.

Similar to \cite{DBLP:journals/pami/WenLLL016}, we define a dense structure on ${\cal G}$ as a sub-hypergraph that has the maximum affinities combining all hyperedges, edges and self-loops of nodes. We introduce an indicator variable ${\bf y}=(y_1,\cdots,y_{n})^{\top}$, such that $\sum_{i=1}^{n}y_{i}=1$, and $y_{i}=\{0, 1/\alpha\}$, where $\alpha$ is the number of nodes in the dense structure. The affinity summation of the hyperedges, edges and self-loops of nodes of the dense structure can be calculated as
\begin{equation}
\begin{array}{ll}
\Theta({\bf y}) = \sum_{d=1}^{D}\lambda_{d}\sum_{{\bf v}_{1:d}\in{\it V}}
{\cal A}({\bf v}_{1:d})\overbrace{y_{1} \cdots y_{d}}^{d}
\end{array}
\label{equ:formulation}
\end{equation}
where ${\bf v}_{1:d}=\{v_{1}, \cdots, v_{d} \}$, $y_{i}$ is the indicator variable corresponding to node $v_i$ ($i=1,\cdots,d$), \ie, $y_{i}=1/\alpha$ if node $v_i$ belongs to the dense structure; otherwise, $y_{i}=0$. Thus, $y_1\cdots y_d$ indicates the confidence of the hyperedge ($d>2$), edge ($d=2$), or self-loop ($d=1$) ${\bf v}_{1:d}$ included in the dense structure. Weights $\lambda_{1},\cdots,\lambda_{D}$ are used to balance the significance of different degrees of hyperedges\footnote{Notably, in this paper, we use the terminology ``affinity'' to indicate the value associated to each edge/hyperedge, which reflects the similarities of the nodes in the corresponding edge/hyperedge. Meanwhile, the terminology ``weight'' is adopted to indicate the numbers used to balance the significance of different degrees of hyperedges, edges and self-loops in dense structure searching. The weights of $d$-th hyperedges may consist of $\kappa>1$ terms (\eg, the weights of the second degree hyperedges may consist of the appearance similarity and motion consistency between two tracklets). In such cases, the weight $\lambda_{d}$ is a vector with the size $1\times\kappa$, and the affinity ${\cal A}({\bf v}_{1:d})$ is also a vector with the size $\kappa\times1$.}. The affinity summation from degree $1$ to $D$ in~\eqref{equ:formulation} describes the overall affinity score combining all the hyperedges, edges, and self-loops of the nodes in the dense structure. Thus, we need to maximize the overall affinity score to exploit the dense structures to complete multi-object tracking.

{\noindent {\bf MOT formulation.}}
We use the non-uniform hypergraph to encode the relations among different tracklets. For each video clip, MOT is initialized by the tracklets\footnote{Our definition of tracklet generalizes cases for single detection, \ie, $m_i=1$, or continuous sequence of detections, \ie, the frame index set $\{ t_1^{i}, \cdots, t_{m_i}^{i}\}$ corresponding the detections on the tracklet, where $t_j^{i}$ is an integer, and $t_j^{i}<t_{j+1}^{i}$, $j=1, \cdots, m_i-1$.}. Let ${\bf T}=\{{\it T}_1, \cdots, {\it T}_n \}$ be the tracklet set in the video sequence, where ${\it T}_{i}$ is the $i$-th tracklet. ${\it T}_i=\{ {\it B}_1^{i}, \cdots, {\it B}_{m_i}^{i}\}$ consists of $m_i$ frame detections, and ${\it B}_{j}^{i}=(x_j^i, y_j^i, w_j^i, h_j^i, t_j^i)$, where $(x_j^i, y_j^i)$ and $(w_j^i, h_j^i)$ are center location and dimension of the detection, and $t_j^i$ is the corresponding frame index.

We formulate the MOT problem as searching dense structures on a non-uniform hypergraph ${\cal G}({\it V}, {\cal E}, {\cal A})$\footnote{Specifically, we only consider the edges/hyperedges with no duplicate nodes, \ie, each edge/hyperedge contains different nodes.}. We set every node in ${\cal G}$ as the starting point, and search the corresponding dense structure from their neighborhoods. Specifically, for a starting point $v_s$, we initialize the indicator variable $y_i^\circ=\frac{1}{|{\cal N}(v_s)|}$, $i=1,\cdots,|{\cal N}(v_s)|$, where $|{\cal N}(v_s)|$ is the number of nodes in $v_s$'s neighborhood. For node $v_s$, the dense structure searching problem is formulated as
\begin{equation}
\begin{array}{ll}
&\argmax_{{\bf y}}\sum_{d=1}^{D}\lambda_{d}\sum_{{\bf v}_{1:d}\in{\cal N}(v_s)}
{\cal A}({\bf v}_{1:d})\overbrace{y_{1} \cdots y_{d}}^{d}  \\
&\text{s.t.} \ \sum_{i=1}^{|{\cal N}(v_s)|} y_{i} = 1, \ y_{s}=\frac{1}{\alpha}, \ \forall{i}, \ y_i\in\{0, \frac{1}{\alpha} \},
\end{array}
\label{equ:dense-structure-search}
\end{equation}
where ${\cal N}(v_s)$ is the neighborhood of node $v_s$. Notably, the constraint $y_s=1/\alpha$ indicates that the node $v_s$ is included in the searched dense structure, and $y_i=1/\alpha$ indicates that the $i$-th node in ${\cal N}(v_s)$ is included in the searched dense structure, otherwise, $y_i=0$.

The problem in \eqref{equ:dense-structure-search} is a combinational optimization problem, since we cannot know the number of nodes in the dense structure $\alpha$ priorly. To reduce the complexity of this NP-hard problem, we relax the constraint $y_i\in\{0, \frac{1}{\alpha}\}$ to $y_i\in[0, \frac{1}{\alpha}]$. In addition, we set a minimal size of the sub-hypergraph to be a constant number $\hat{\alpha}$ to avoid the degeneracy, \ie, $\hat{\alpha}\leq\alpha$. Thus, the constraint is converted to $y_i\in[0, \frac{1}{\hat{\alpha}}]$. We would like to highlight that the objective function for dense structure exploiting in \cite{DBLP:journals/pami/WenLLL016} is a specific case of \eqref{equ:dense-structure-search}, \ie, if we set $\lambda_{d^\star} \neq {\bf 0}$ for a specific $d^\star\geq3$, and make $\lambda_{d}=0$, $\forall d\ne{d}^\star$, the non-uniform hypergraph ${\cal G}$ will degenerate into a $d^\star$-uniform hypergraph, and the objective in \eqref{equ:dense-structure-search} becomes similarly to that in \cite{DBLP:journals/pami/WenLLL016}. The optimization algorithm in \cite{DBLP:journals/pami/WenLLL016} for uniform hypergraph model cannot be directly applied to solve the problem in \eqref{equ:dense-structure-search}.

After exploiting the dense structures, the radical post-processing strategy presented in \cite{DBLP:conf/cvpr/Longyin14} is adopted to remove the conflicts among the searched dense structures. Then, we stitch the tracklets in each post-processed dense structures to form the long trajectories.


{\noindent {\bf Enforcing edge/hyperedge constraints.}}
In the practical MOT scenarios, the objects have two physical constraints: 1) one object cannot occupy two different places at a time; 2) the velocity of a object is below certain maximum possible velocity. As such, in constructing the hypergraph, two nodes connected by one edge/hyperedge should not overlap in time, and the distance between the last and first detections of the tracklet should not larger than the maximal distance that can reach with the maximal possible velocity. These two constraints can reduce the number of edges and hyperedges and computational complexity.

{\noindent {\bf Calculating self-loop affinity.}}
We associate a node with a score to reflect its reliability being a true tracklet of an object, \ie, ${\cal A}(v_i) = \rho(v_i)$, where $\rho(v_i)$ ($0\leq\rho(v_i)\leq1$) is the confident score of the tracklet $v_i$ calculated by averaging the scores of all detections in the tracklet.

{\noindent {\bf Calculating edge affinity.}}
The edges in the hypergraph encode the similarities between two nodes (tracklets), which consists of three terms: HSV histogram similarity ${\cal P}_{\text{col}}(v_i,v_j)$, CNN feature similarity ${\cal P}_{\text{cnn}}(v_i,v_j)$, and local motion similarity ${\cal P}_{\text{mot}}(v_i,v_j)$, \ie, ${\cal A}(v_i,v_j) = \big[{\cal P}_{\text{col}}(v_i,v_j), {\cal P}_{\text{cnn}}(v_i,v_j), {\cal P}_{\text{mot}}(v_i,v_j) \big]$.

Specifically, the HSV histogram similarity ${\cal P}_{\text{col}}(v_i,v_j)$ is calculated as ${\cal P}_{\text{col}}(v_i,v_j)= \chi\big(h^{-}(v_i), h^{+}(v_j)\big)$, where $\chi(\cdot,\cdot)$ is the cosine similarity between the HSV histograms of the detections in the last frame of $v_i$ (\ie, $h^{-}(v_i)$) and the first frame of $v_j$ (\ie, $h^{+}(v_j)$).

Moreover, the CNN feature similarity ${\cal P}_{\text{cnn}}(v_i,v_j)$ is calculated as
${\cal P}_{\text{cnn}}(v_i,v_j) = \frac{1 + \chi\big( \mu^{-}(v_i), \mu^{+}(v_j)\big)}{2}$,
where $\mu^{-}(v_i)$ and $\mu^{+}(v_j)$ are the CNN features of the detections in the last frame of $v_i$ and the first frame of $v_j$.

Finally, the similarity between two bounding boxes based on the generalized KLT tracker \cite{DBLP:conf/cvpr/ZhouTW13} is calculated as
${\cal P}_{\text{mot}}(v_i,v_j) = 1 - \frac{2}{1+\exp\Big(\frac{2\cdot\zeta(v_i,v_j)}{\gamma(B^i_{m_i})+\gamma(B^j_{1})}\Big)}$,
where $\gamma(B^i_{m_i})$ and $\gamma(B^j_{1})$ are the areas of the detections in the last frame of $v_i$ and the first frame of $v_j$, and $\zeta(v_i,v_j)$ is the number of point trajectories generated by KLT tracker across the bounding boxes of both the first frame of $v_i$ and first frame of $v_j$.

{\noindent {\bf Calculating hyperedge affinity.}}
We count the number of local point trajectories passing through the regions of ${\bf v}_{1:d}$ to calculate the affinities of hyperedges, which encodes the motion consistency of tracklets ${\bf v}_{1:d}$. Thus, for the $i$-th hyperedge with degree $d$, the affinity is calculated as ${\cal A}({\bf v}_{1:d})= 1 - \frac{2}{1+\exp\Big(\frac{d\cdot\zeta({\bf v}_{1:d})}{\sum_{u=1}^{d}\sum_{j=1}^{l_u}\gamma(B^{u}_{j})}\Big)}$, where $\zeta({\bf v}_{1:d})$ measures the number of local point trajectories crossing all regions of ${\bf v}_{1:d}$, $l_u$ is the length of tracklet $v_u$, $B^{u}_j$ is the $j$-th detection on $v_u$, and $\gamma(B^{u}_{j})$ is the area of the detection $B^{u}_j$.

{\noindent \textbf{Near-online tracking.}}
It is difficult to handle all detections in a long video sequences at a time, since it requires large memory and computation sources to construct non-uniform hypergraphs and perform dense structure search on all detections. In order to achieve both accuracy and efficiency, inspired by \cite{DBLP:conf/iccv/Choi15}, we use a near-online strategy for MOT. Specifically, after getting $\tau$ video frames at time $t$, we construct a {\em non-uniform} hypyergraph to describe the hybrid orders of dependencies among detections and search the dense structures on the hypergraph to generate short tracklets in the temporal window $[t-\tau, t]$. Then, we construct a conventional graph\footnote{The conventional graph is a special case of the non-uniform hypergraph, which only includes the conventional edges in the graph, \ie,  ${\it D}=2$.} to describe the associations between the tracked targets and the short tracklets within $[t-\tau, t]$. After that, we perform the dense structure searching on the conventional graph to associate the short tracklets and the tracked targets to get the final trajectories at the current time stamp. This process is carried out repeatedly every $\tau$ frames to complete the tracking task in the whole video.

\section{Inference}
For efficiency, we use the simple pairwise update algorithm \cite{DBLP:journals/ijcv/LiuYLY12} to solve the dense structure searching problem on hypergraph ${\cal G}$ corresponding to node $v_s$ in \eqref{equ:dense-structure-search}. We first form the Lagrangian of the problem as
\begin{equation}
\begin{array}{ll}
&{\cal L}({\bf y}, a, {\bf b}, {\bf c}) = \Theta({\bf y}) - a \cdot \big( \sum_{i=1}^{|{\cal N}(v_s)|} y_i - 1 \big) \\
&+ \sum_{i, i\ne{v_s}} b_i\cdot y_{i} + \sum_{i, i\ne{v_s}} c_{i} \cdot (\frac{1}{\hat{\alpha}} - y_{i}),
\end{array}
\label{equ:lagrangian}
\end{equation}
where $a$, ${\bf b}=(b_1,\cdots,b_{|{\cal N}(v_s)|})$, and ${\bf c}=(c_1,\cdots,c_{|{\cal N}(v_s)|})$ are Lagrangian multipliers with $a\ge0$, $b_i\ge0$, and $c_i\ge0$, $i=1,\cdots,|{\cal N}(v_s)|$. Any local maximizer ${\bf y}^\ast$ of the objective function must satisfy the Karush-Kuhn-Tucker (KKT) conditions \cite{DBLP:conf/BS/KKT1951}, \ie,
\begin{equation}
\left\{
\begin{array}{cl}
&\frac{\partial \Theta({\bf y}^\ast)}{\partial y_i} - a + b_i - c_i = 0, i \neq v_s; \\
&\sum_{i, i \neq v_s} y^\ast_i \cdot b_i = 0; \\
&\sum_{i, i \neq v_s} c_i \cdot (\frac{1}{\hat{\alpha}} - y^\ast_i) = 0;\\
&a\ge0, b_i\ge0, c_i\ge0, i=1,\cdots,|{\cal N}(v_s)|;\\
&\sum_{i=1}^{n}y_{i}=1, y_{i}\in[0, \frac{1}{\hat{\alpha}}].
\end{array}
\right.
\label{equ:derivation}
\end{equation}
We define $\phi_{i}({\bf y})=\frac{\partial \Theta({\bf y})}{\partial y_i}$ as {\em reward} at node $v_i$, which is calculated as
\begin{align}
\begin{array}{ll}
\phi_{i}({\bf y}) &= \lambda_{1}{\cal A}(i)\nonumber\\
&+\sum_{d=2}^{D}\lambda_{d}\sum_{{\bf v}_{1:d-1}\in{{\cal N}(v_s)}}{\cal A}({\bf v}_{1:d-1}, i)\prod_{j=1}^{d-1}y_{v_j}.
\end{array}
\end{align}

Since $\forall i$, $y^\ast_i\ge0$, $b_i \ge 0$, $\sum_{i, i\neq{v_s}} y^\ast_i \cdot b_i = 0$, we have that if $y^\ast_i > 0$, then $b_i = 0$. Meanwhile, since $\forall i$, $c_i \ge 0$, and $y^\ast_i \leq \frac{1}{\hat{\alpha}}$, we have that if $0< y^\ast_i < \frac{1}{\hat{\alpha}}$, then $c_i = 0$. In this way, for node $i\neq v_s$, the KKT conditions can be further rewritten as
\begin{equation}
\phi_{i}({\bf y}) = \left \{
\begin{array}{cl}
&\leq a, \quad y^\ast_i = 0, i \neq v_s; \\
&= a, \quad 0 < y^\ast_i < \frac{1}{\hat{\alpha}}, i \neq v_s; \\
&\geq a, \quad y^\ast_i = \frac{1}{\hat{\alpha}}, i \neq v_s.
\end{array}
\right.
\label{equ:kkt-constraints}
\end{equation}
Based on ${\bf y}$ and $\alpha$, we can partition the solution space into three disjoint subsets, $\Omega_{1}({\bf y})=\{i|y_i=0\}$, $\Omega_{2}({\bf y})=\{i|y_i\in(0,\frac{1}{\hat{\alpha}})\}$, and $\Omega_{3}({\bf y})=\{i|y_i = \frac{1}{\hat{\alpha}}\}$. Thus, similar to Theorem 1 in \cite{DBLP:journals/ijcv/LiuYLY12}, we find that there exists an appropriate $a$, such that (1) the rewards at all node in $\Omega_{1}({\bf y})$ are no larger than $a$; (2) the rewards at all nodes in $\Omega_{2}({\bf y})$ are equal to $a$; and (3) the rewards at all nodes in $\Omega_{3}({\bf y})$ are larger than $a$.

A simple pairwise updating method is used to optimize \eqref{equ:dense-structure-search}. That is, we can increase one component $y_p$ and decrease another one $y_q$ appropriately, to increase the objective $\Theta({\bf y})$. To be specific, we first introduce another variable $y_{l}^\prime$ that is defined as: $y_{l}^\prime=y_l$, for $l \neq p$ and $l\neq q$; $y_{l}^\prime=y_l+\eta$, for $l=p$; and $y_{l}^\prime=y_l-\eta$, for $l=q$,
where ${\bf y}^\prime=(y_1^\prime, \cdots, y_{|{\cal N}(v_s)|}^\prime)$ is the updated indicator variable in optimization process. Then, the change of objective after updating is
\begin{align}
\Delta\Theta({\bf y})&=\Theta({\bf y}^\prime) - \Theta({\bf y})\nonumber\\
&=\varphi_{p,q}({\bf y}) \cdot \eta^2 + \big(\phi_{p}({\bf y}) - \phi_{q}({\bf y})\big) \cdot \eta,
\label{equ:update-objective}
\end{align}
where $\varphi_{p,q}({\bf y}) = -\lambda_2\cdot{\cal A}(p,q)-\sum_{d=3}^{D}\lambda_{d}\sum_{{\bf v}_{1:d-2}\ne{p,q}} {\cal A}({\bf v}_{1:d-2},p,q)\prod_{j=1}^{d-2} y_{v_j}$.

To maximize the objective difference $\Delta \Theta({\bf y})$, we select the updating step $\eta$ as follows\footnote{In general, we can assume $\phi_{p}({\bf y})>\phi_{q}({\bf y})$. When $\phi_{p}({\bf y})<\phi_{q}({\bf y})$, we can exchange indexes $p$ and $q$ to maximize $\Delta \Theta({\bf y})$. Please see the supplementary material for more details.}:
\begin{equation}
\small
\eta=\left \{
\begin{array}{cl}
&\min(y_{q},\frac{1}{\hat{\alpha}}-y_{p}), \ \text{if} \ \varphi_{p,q}({\bf y})\ge0; \\
&\min\big(y_{q},\frac{1}{\hat{\alpha}}-y_{p},\frac{\phi_{q}({\bf y})-\phi_{p}({\bf y})}{2\cdot\varphi_{p,q}({\bf y})}\big), \ \text{if} \ \varphi_{p,q}({\bf y})<0; \\
&\min(y_{q},\frac{1}{\hat{\alpha}}-y_{p}), \ \text{if} \ \phi_{p}({\bf y})=\phi_{q}({\bf y}), \varphi_{p,q}({\bf y}) > 0.
\end{array}
\right.
\label{equ:update-step}
\end{equation}

We use a heuristic strategy to compute a local maximizer ${\bf y}^\ast$ of \eqref{equ:dense-structure-search}, \ie, gradually select pairs of nodes $(v_{p}, v_{q})$ to maximize the increase of $\Theta({\bf y})$ by updating the indicator variable ${\bf y}$ based on the updating step $\eta$ calculated by \eqref{equ:update-step}. Specifically, from \eqref{equ:update-objective} and \eqref{equ:update-step}, we find that (1) if $\phi_{p}({\bf y})>\phi_{q}({\bf y})$, there exists $a$ such that the objective $\Theta({\bf y})$ can be increased by updating ${\bf y}$ based on \eqref{equ:update-objective}; (2) when $\phi_{p}({\bf y})=\phi_{q}({\bf y})$ and $\varphi_{p,q}({\bf y})>0$, the objective $\Theta({\bf y})$ can be increased by increasing either $y_{p}$ or $y_{q}$, and decreasing the other one; (3) when $\phi_{p}({\bf y})=\phi_{q}({\bf y})$ and $\varphi_{p,q}({\bf y})=0$, the objective $\Theta({\bf y})$ will not be affected by changing ${\bf y}$.

Thus, in each iteration, we can select node $v_{p}$ with the largest reward from set $\Omega_{1}\cup\Omega_{2}$, \ie, $v_{p}\in\Omega_{1}\cup\Omega_{2}$, and node $v_{q}$ with the smallest reward from set $\Omega_{2}\cup\Omega_{3}$, \ie, $v_{q}\in\Omega_{2}\cup\Omega_{3}$, satisfying $\phi_{p}({\bf y})>\phi_{q}({\bf y})$, to increase $\Theta({\bf y})$ by increasing $y_{p}$ and decreasing $y_{q}$ with an appropriate $\eta$ in \eqref{equ:update-step}. This process is iterated until the reward of $v_{p}$ equals to $v_{q}$. If $\Theta({\bf y})$ can not be increased according to \eqref{equ:update-objective}, then ${\bf y}$ is already a local maximizer. The overall procedure is summarized in Algorithm \ref{alg:pairwise-update}.

\begin{algorithm}[t]
\caption{Compute the local maximizer ${\bf y}^\ast$}
\label{alg:pairwise-update}
\footnotesize{
\begin{algorithmic}[1]
\REQUIRE The affinity set ${\cal A}$ corresponding to the hyperedges in ${\cal G}$, the starting point ${\bf y}^\circ=(y_{1}^\circ,\cdots,y_{|{\cal N}(v_s)|}^\circ)$ and the minimal size of sub-hypergraph $\hat{\alpha}$.
\STATE Initialize the indicator variable ${\bf y}={\bf y}^\circ$.
\WHILE{${\bf y}$ is the local maximizer}
\STATE Select $v_{p}\in\Omega_{1}\cup\Omega_{2}$ with the largest reward $\phi_{p}({\bf y})$;
\STATE Select $v_{q}\in\Omega_{2}\cup\Omega_{3}$ with the smallest reward $\phi_{q}({\bf y})$;
\IF{$\phi_{p}({\bf y})>\phi_{q}({\bf y})$}
\STATE Compute $\eta$ according to \eqref{equ:update-step}, update ${\bf y}$ and the corresponding rewards.
\ELSIF{$\phi_{p}({\bf y})=\phi_{q}({\bf y})$}
\STATE Find another pair of nodes $(v_i, v_j)$ satisfying $\varphi_{i,j}({\bf y})>0$ and $\phi_{i}({\bf y})=\phi_{j}({\bf y})$, where $v_{i}\in\Omega_{1}\cup\Omega_{2}$ and $v_{j}\in\Omega_{2}\cup\Omega_{3}$.
\IF{such a pair exists}
\STATE Compute the corresponding $\eta$ according to \eqref{equ:update-step}.
\STATE Update ${\bf y}$ and the corresponding rewards.
\ELSE
\STATE ${\bf y}$ is already a local maximizer, \ie, ${\bf y}^\ast={\bf y}$.
\ENDIF
\ENDIF
\ENDWHILE
\ENSURE The local maximizer indicator variable ${\bf y}^\ast$.
\end{algorithmic}}
\end{algorithm}
\vspace{-2mm}

\subsection{Learning}
\label{sec_learn_confidence}
Instead of selecting the weights $\bm{\lambda}=(\lambda_1,\cdots,\lambda_D)$ in \eqref{equ:formulation} empirically, we use a structured SVM \cite{DBLP:journals/ml/JoachimsFY09} to learn $\bm{\lambda}$ automatically from the training data. Specifically, given a set of ground-truth bounding boxes of objects in the $j$-th training video ($1\leq{j}\leq{U}$, where $U$ is the total number of training videos), we aim to recover the trajectories of objects, which is equivalent to cluster the input bounding boxes into several groups. That is to obtain the indicator variables of the clusters ${\bf Y}_{j}=({\bf y}_{1,j},\cdots,{\bf y}_{k_j,j})$, where ${\bf y}_{i,j}$ ($1\leq{i}\leq{k_j}$) is the indicator variable of the $i$-th target, and $k_j$ is the total number of targets in the video. The bounding boxes in each group belong to the same target.

The function defined in \eqref{equ:formulation} can be rewritten as a linear function of $\bm{\lambda}$, \ie, $\Theta({\bf Y}_{j})= \bm{\lambda}^{\top} \cdot {\cal S}({\bf Y}_{j})$, where
\begin{align}
\begin{array}{ll}
{\cal S}({\bf Y}_{j}) &= \big[ \sum_{\varsigma=1}^{k_j}\sum_{v_i\in{\it V}}{\cal A}(v_i)y_{\varsigma,i}, \cdots,\nonumber\\
 &\sum_{\varsigma=1}^{k_j}\sum_{{\bf v}_{1:D}\in{\it V}} {\cal A}({\bf v}_{1:D})\prod_{i=1}^{D}y_{\varsigma,i} \big].
 \end{array}
\end{align}
We aim to find the optimal weights $\bm{\lambda}$ by maximizing the objective function $\Theta({\bf Y}_{j})$ with the same input object detections. Then, the objective using a SSVM with margin rescaling is formulated as
\begin{equation}
\begin{array}{ll}
&\min_{\bm{\lambda}} {\frac{1}{2}\|\bm{\lambda}\|_{2}}+C\cdot\sum_{j=1}^{U} \xi_j, \\
\textrm{s.t.}\ &\bm{\lambda}^{\top}\Big({\cal S}({\bf Y}^\ast_{j})-{\cal S}({\bf Y}_{j})\Big)+\xi_j \geq \Delta({\bf Y}_{j},{\bf Y}^\ast_{j}),\\
& \xi_j \geq 0, \ \ j = 1,\cdots,U.
\end{array}
\label{equ:ssvm}
\end{equation}

Intuitively, this formulation requires that the score $\bm{\lambda}^{\top}\cdot{\cal S}({\bf Y}^\ast_{j})$ of any ground-truth annotated video must be larger than the score $\bm{\lambda}^{\top}\cdot{\cal S}({\bf Y}_{j})$ of any other results ${\bf Y}_{j}$ by the loss $\Delta({\bf Y}_{j},{\bf Y}^\ast_{j})$ minus the slack variable $\xi_j$. The constant $C$ adjusts the importance of minimizing the slack variables. The loss function $\Delta({\bf Y}_{j},{\bf Y}^\ast_{j})$ measures how incorrect ${\bf Y}_{j}$ is according to the weighted Hamming loss in \cite{DBLP:conf/bmvc/WangF15}. Meanwhile, the SSVM formulation in \eqref{equ:ssvm} has exponential number of constraints for each training sequence. We use a cutting plain algorithm \cite{DBLP:journals/ml/JoachimsFY09} to solve this problem, which has time complexity linear in the number of training examples.

\section{Experiments}
We conduct experiments on several popular MOT evaluation datasets, \ie, the multi-pedestrian tracking \cite{DBLP:journals/pami/WenLLL016} (including the PETS09 and ParkingLot sequences), MOT2016 \cite{DBLP:journals/corr/MilanL0RS16}, and multi-face tracking \cite{DBLP:journals/pami/WenLLL016} datasets, to evaluate the performance of the proposed MOT method (denoted as NT subsequently)\footnote{The source code of the proposed method is available at \url{https://github.com/longyin880815}.}. We use the MOT2016-train set to train the set-to-set recognition model \cite{DBLP:conf/cvpr/LiuYO17} to calculate the CNN feature similarity, and the multi-pedestrian tracking dataset to analyze the influence of the degree of hypergraph to tracking performance. In addition, we conduct the ablation study to demonstrate the effectiveness of non-uniform hypergraph and SSVM learning.

{\noindent \textbf{Evaluation Metrics.}}
Following previous MOT methods, we use the widely adopted multi-object tracking accuracy (MOTA) metric~\cite{bernardin2008evaluating} to compare the performance of the trackers. MOTA is a cumulative measure combing false negatives (FN), false positives (FP), and identity switches (IDS). We report mostly tracked (MT), mostly lost (ML), FP, FN, IDS, and the fragmentation of the tracked objects (FM) to measure a tracker comprehensively. In addition, for the multi-pedestrian and multi-face tracking datasets~\cite{DBLP:journals/pami/WenLLL016}, we also report the multi-object tracking precision (MOTP) score, which computes the total error of tracked positions comparing with the manually annotated ground-truth, with normalization to the hit/miss threshold value. Following the evaluation protocol in MOT2016, we use the ID F1 score (IDF1) \cite{DBLP:conf/eccv/RistaniSZCT16} instead of MOTP, which is the ratio of correctly identified detections over the average number of ground-truth and computed detections.

{\noindent \textbf{Parameters.}}
We conduct an experiment to select the maximal degree of the hypergraph ${\it D}$. We set ${\it D}=2, \cdots, 5$ while keeping other parameters fixed, and denote the resulting models as NT\_d(2), $\cdots$, NT\_d(5). For each maximal degree, we use the sequences in the training set of MOT2016 to learn the weights of different degrees of hyperedges $\bm{\lambda}=(\lambda_1,\cdots,\lambda_{D})$ using SSVM, and use the sequences in multi-pedestrian tracking dataset for testing. The uniform average performance of the trackers in multi-pedestrian tracking dataset is presented in Table \ref{tab:nt-variants}. Specifically, we divide each sequence in the MOT2016 train-set into non-overlapping sequences of $14$ frames. And then, we take the detections that have more than $50\%$ overlap with the ground-truth as true detections to collect training samples for the weights $\bm{\lambda}$ learning.

As shown in Table \ref{tab:nt-variants}, NT achieves the best performance with the maximal degree ${\it D}=4$, indicated by higher MOTA and lower IDS and FM scores. We notice that the performance of NT decreases when ${\it D}>4$, this may be because the hypergraph with excessive high degree fails to describe the motion patterns of objects well, particularly for the objects moving fast with drastic variations of directions. Thus, we set ${\it D}=4$ in our experiments, and the learned weights of different degree of hyperedge are $\lambda_1=0.58535$, $\lambda_2=[0.15576,3.0332,0.34388]$, $\lambda_3=1.2879$, and $\lambda_4=0.22324$. The batch size $\tau$ in near-online tracking is set to $7$. The minimal size of the sub-hypergraph is set as $\hat{\alpha}=2$. We fix all parameters to these values in the experiments.

\begin{table}[t]
\centering
\caption{Comparisons of variants of the proposed NT tracker on multi-pedestrian tracking dataset.}
\footnotesize \setlength{\tabcolsep}{1.6pt}
\begin{tabular}{c|c|c|cccc}
\hline
Variants &$D$              &$\bm{\lambda}$                                 &MOTA      &MOTP       &IDS      &FM   \\ \hline
NT\_d(2)      &$2$              &learned                                 &67.5	      &62.4	      &103.7	&92.2  \\ \hline
NT\_d(3)      &$3$              &learned                                 &68.8	      &64.5	      &83.8	    &76.2  \\ \hline
NT\_d(4)      &$4$              &learned                                 &{\bf 68.9} &{\bf 65.0}     &68.3 &68.8  \\ \hline
NT\_d(5)      &$5$              &learned                                 &68.5      &64.7         &\textbf{61.5}      &\textbf{63.7}  \\ \hline
\hline
NT\_r(4)      &$4$              &learned,  $\lambda_i=0$, $i=3$     &68.4 &63.5 &72.7 &74.2  \\ \hline
NT\_r(5)      &$5$              &learned,  $\lambda_i=0$, $i=3,4$    &67.6 &63.5 &64.3 &66.0  \\ \hline
\hline
NT\_e(2)      &$2$              &$\lambda_i=1$, $i=1,2$                  &67.1 &62.6 &103.7 &87.0  \\ \hline
NT\_e(3)      &$3$              &$\lambda_i=1$, $i=1,\cdots,3$           &67.5 &63.7 &103.3 &87.5  \\ \hline
NT\_e(4)      &$4$              &$\lambda_i=1$, $i=1,\cdots,4$           &67.4 &63.7 &104.0 &86.7  \\ \hline
NT\_e(5)      &$5$              &$\lambda_i=1$, $i=1,\cdots,5$           &67.1 &64.6 &93.2  &81.7  \\ \hline
\end{tabular}
\label{tab:nt-variants}
\vspace{-1mm}
\end{table}

{\noindent \textbf{Ablation Study.}}
To demonstrate the contribution of non-uniform hypergraph, we construct two variants of the proposed NT tracker by removing the hyperedges with certain degrees, \ie, NT\_r(3) and NT\_r(4), and evaluate them on the multi-pedestrian tracking dataset \cite{DBLP:journals/pami/WenLLL016}, shown in Table \ref{tab:nt-variants}. The results in Table \ref{tab:nt-variants} shows that removing the hyperedges with degrees $3$ and $4$ will negatively affect the performance (\ie, reduce $0.5\%$ and $0.9\%$ MOTA scores), which shows that exploiting different degrees of dependencies among objects is important for MOT performance.

Besides, to demonstrate the contribution of SSVM, in Table \ref{tab:nt-variants}, we present the performance of non-uniform hypergraph based trackers with equal weights of different degrees of hyperedges in multi-pedestrian tracking, denoted as NT\_e(2), $\cdots$, NT\_e(5). The NT\_d($i$) methods perform consistently better than the NT\_e($i$) methods with the same maximal degrees, \eg, NT\_d(2) {\it vs.} NT\_e(2), and NT\_d(5) {\it vs.} NT\_e(5), where $i=2,\cdots,5$. The results show that using SSVM to learn the weights of hyperedges of different degrees can improve the performance.

{\noindent \textbf{Multi-Pedestrian Tracking.}}
We perform experiments for the multi-pedestrian tracking on five sequences from the PETS09 dataset \cite{DBLP:conf/avss/EllisF10}: S2L1 ($795$ frames), S2L2 ($436$ frames), S2L3 ($240$ frames), S1L1-1 ($221$ frames), and S1L1-2 ($241$ frames), and ParkingLot sequence from \cite{DBLP:conf/eccv/ZamirDS12} ($996$ frames). These sequences are captured in the crowded surveillance scenes with frequent occlusions, abrupt motion, illumination changes, etc. Following \cite{DBLP:journals/pami/WenLLL016,DBLP:conf/cvpr/AndriyenkoSR12}, we report the uniform average scores on different metrics over sequences of the proposed NT algorithm, as well as five state-of-the-art trackers, \ie, KSP \cite{DBLP:journals/pami/BerclazFTF11}, DPMF \cite{DBLP:conf/cvpr/PirsiavashRF11}, CEM \cite{DBLP:conf/cvpr/AndriyenkoS11}, DCT \cite{DBLP:conf/cvpr/AndriyenkoSR12} and FH\textsuperscript{2}T \cite{DBLP:journals/pami/WenLLL016}, in Table \ref{tab:pedestrian-results}. The tracking results of previous methods are taken from \cite{DBLP:journals/pami/WenLLL016}. For fair and comprehensive comparisons, we use the same frame detections, ground-truth annotations as well as the evaluation protocol provided by the authors of \cite{DBLP:journals/pami/WenLLL016}. We train the set-to-set recognition method \cite{DBLP:conf/cvpr/LiuYO17} based on the pre-trained GoogLeNet \cite{DBLP:conf/cvpr/SzegedyLJSRAEVR15} on the training set of MOT2016 to extract the CNN features of the detections.

\begin{table}[t]
\caption{Comparison of the proposed tracker with the previous trackers in multi-pedestrian tracking sequences.}
  \label{tab:pedestrian-results}
\centering
\footnotesize \setlength{\tabcolsep}{1.5pt}
\begin{tabular}{c|c|c|c|c|c|c|c|c}
\hline
Method &MOTA &MOTP &MT[\%] &ML[\%] &FP &FN &IDS &FM \\
\hline
\hline
KSP &45.5 &67.1 &33.4 &35.6 &107.8 &2223.2 &{\bf 42.2} &{\bf 49.8} \\
DPMF  &51.6 &{\bf 70.0} &21.5 &27.0 &{\bf 68.8}  &1897.0 &61.8 &80.7 \\
CEM    &55.7 &66.6 &30.1 &21.7 &127.3 &1652.8 &63.7 &56.7 \\
DCT   &58.1 &67.6 &43.1 &21.3 &119.5 &1610.2 &64.2 &53.2 \\
FH\textsuperscript{2}T   &66.2 &64.9 &54.3 &14.7 &194.5 &1150.8 &45.2 &73.7 \\
\hline
NT  &{\bf 68.9} &65.0 &{\bf 58.2} &{\bf 9.6} &252.7 &{\bf 974.3} &68.3 &68.8\\
\hline
\end{tabular}
\vspace{-1mm}
\end{table}

As shown in Table \ref{tab:pedestrian-results}, we find that our NT tracker performs better than the state-of-the-art methods on several important metrics (\eg, MOTA, MT, and ML). Specifically, NT improves $2.7\%$ and $3.9\%$ average MOTA and MT scores, and reduces $5.1\%$ average ML score, against the second best tracker FH\textsuperscript{2}T \cite{DBLP:journals/pami/WenLLL016}. This may be attributed to that our method uses non-uniform hypergraph instead of uniform hypergraph in \cite{DBLP:journals/pami/WenLLL016}, especially for tracking in crowded scenes with different motions and frequent occlusions of objects. By the way, we notice that the FH\textsuperscript{2}T method \cite{DBLP:journals/pami/WenLLL016} performs better than the methods (\eg, DPMF \cite{DBLP:conf/cvpr/PirsiavashRF11} and DCT \cite{DBLP:conf/cvpr/AndriyenkoSR12}), both only considering the similarities between pairs of tracklets (\ie, FH\textsuperscript{2}T \cite{DBLP:journals/pami/WenLLL016} produces $14.6\%$ and $8.1\%$ higher average MOTA score than DPMF \cite{DBLP:conf/cvpr/PirsiavashRF11} and DCT \cite{DBLP:conf/cvpr/AndriyenkoSR12}), which indicates that exploiting the high-order similarities among multiple tracklets is crucial for MOT.

{\noindent \textbf{MOT2016 Benchmark.}}
The MOT2016 benchmark \cite{DBLP:journals/corr/MilanL0RS16} is a collection of $14$ video sequences ($7$/$7$ for training and testing, respectively), with a relatively high variations in object movements, camera motion, viewing angle and crowd density. The benchmark primarily focuses on pedestrian tracking. The ground-truths for testing set are strictly invisible to all methods, \ie, all results on testing set were submitted to the respective testing servers for evaluation. We use the training set to learn the parameters of the proposed algorithm, and submit our results on testing set for evaluation, shown in Table \ref{tab:mot16-test-results}. For a fair comparison with the state-of-the-art MOT methods, we use the reference object detections provided by the benchmark \cite{DBLP:journals/corr/MilanL0RS16}. We train the set to set recognition method \cite{DBLP:conf/cvpr/LiuYO17} based on the pre-trained GoogLeNet \cite{DBLP:conf/cvpr/SzegedyLJSRAEVR15} on the training set of MOT2016 to extract the CNN features of the detections.

\begin{table}[t]
\caption{Comparison of the proposed tracker with the state-of-the-art trackers in the test set of the MOT2016 benchmark (accessed on 08/18/2018).}
\label{tab:mot16-test-results}
\centering
\footnotesize \setlength{\tabcolsep}{0.1pt}
\begin{tabular}{c|ccccccccc}
\hline
Method &MOTA &IDF1 &MT[\%] &ML[\%] &FP &FN &IDS &FM &Hz\\
\hline
\multicolumn{10}{c}{\textit{online:}}\\
\hline
EAMTT   &38.8 &42.4 &7.9  &49.1 &8,114	 &102,452 &965   &1,657 &{\bf 11.8}\\
DCCRF &44.8 &39.7 &14.1 &42.3 &5,613 &94,133 &968 &1,378 &0.1\\
STAM  &46.0 &50.0 &14.6 &43.6 &6,895	 &91,117 &473 &1,422 &0.2\\
AMIR    &47.2 &46.3 &14.0 &41.6  &2,681 &92,856  &774 &1,675 &1.0\\
\hline\hline
\multicolumn{10}{c}{\textit{offline:}}\\
\hline
Quad   &44.1 &38.3 &14.6 &44.9 &6,388 &94,775 &745 &1,096 &1.8\\
INT &45.4 &37.7 &18.1 &38.7 &13,407 &85,547 &600 &930 &4.3\\
MHT    &45.8 &46.1 &16.2 &43.2 &6,412  &91,758  &590  &781  &0.8\\
NLPa &47.6 &47.3 &17.0 &40.4 &5,844 &89,093 &629 &768 &8.3\\	
FWT    &47.8	&44.3 &19.1 &38.2 &8,886 &85,487 &852 &1,534 &0.6\\
LMP    &{\bf 48.8}	&51.3 &18.2 &40.1 &6,654 &86,245 &481 &595 &0.5\\
\hline
\hline
\multicolumn{10}{c}{\textit{near-online:}} \\
\hline
NOMT   &46.4 &{\bf 53.3} &18.3 &41.4 &9,753  &87,565  &{\bf 359}   &{\bf 504} &2.6\\
Ours     &47.5 &43.6 &{\bf 19.4} &{\bf 36.9} &13,002  &{\bf 81,762} &1,035 &1,408 &0.8\\
\hline
\end{tabular}
\end{table}

In Table \ref{tab:mot16-test-results}, NT is compared with the state-of-the-art methods including EAMTT \cite{DBLP:conf/eccv/Sanchez-Matilla16}, Quad \cite{DBLP:conf/cvpr/SonBCH17}, MHT \cite{DBLP:conf/iccv/KimLCR15}, STAM \cite{DBLP:conf/iccv/ChuOLWLY17}, NOMT \cite{DBLP:conf/iccv/Choi15}, AMIR \cite{DBLP:conf/iccv/SadeghianAS17}, NLPa \cite{DBLP:conf/cvpr/EvgencyL17}, FWT \cite{DBLP:journals/corr/HenschelLCR17}, LMP \cite{DBLP:conf/cvpr/SyT17}, INT \cite{DBLP:journals/tip/LanWZTGH18}, and DCCRF \cite{DBLP:journals/corr/abs-1806-01183}. Our NT method performs on par with the state-of-the-art trackers (\eg, FWT and LMP) in terms of tracking accuracy. Specifically, LMP uses additional person re-identification datasets to train a deep StackNet with body part fusion to associate pedestrians across frames, achieving the top tracking accuracy (\ie, $48.8\%$ MOTA), while FWT incorporates multiple detectors to improve the tracking performance. In contrast to the aforementioned methods using complex appearance model, our NT algorithm focuses on exploiting different degrees of dependencies among tracklets to assemble various kinds of appearance and motion patterns. The appearance modeling strategies proposed in those methods are complementary to our NT tracker. Meanwhile, we notice that NT achieves better performance than the high-order information based MHT in terms of tracking accuracy ($47.5\%$ {\em vs.} $45.8\%$), which implies that exploiting adaptive dependencies among objects is important for MOT.

{\noindent \textbf{Multi-Face Tracking.}}
In addition to pedestrian tracking, we also evaluate NT on the SubwayFaces dataset used in \cite{DBLP:journals/pami/WenLLL016}. The dataset consists of four sequences, namely S001, S002, S003, and S004 with $1,199$, $1,000$, $1,600$, and $1,001$ frames, captured from surveillance videos in subway with manually annotations. We compare our approach with five state-of-the-art MOT algorithms, \ie, CEM \cite{DBLP:conf/cvpr/AndriyenkoS11}, KSP \cite{DBLP:journals/pami/BerclazFTF11}, DCT \cite{DBLP:conf/cvpr/AndriyenkoSR12}, DPMF \cite{DBLP:conf/cvpr/PirsiavashRF11} and FH\textsuperscript{2}T~\cite{DBLP:journals/pami/WenLLL016}, with uniform average scores on different metrics over sequences presented in Table \ref{tab:face-results}. We use the same input detections, ground-truth annotations and the evaluation protocol as \cite{DBLP:journals/pami/WenLLL016}, and the results of the state-of-the-art trackers in Table \ref{tab:face-results} are taken from \cite{DBLP:journals/pami/WenLLL016}. We use pre-trained AlexNet \cite{DBLP:conf/nips/KrizhevskySH12} to extract the CNN features of the detected faces.

\begin{table}[t]
\caption{Comparison of the proposed tracker with other state-of-the-art trackers in the SubwayFace dataset.}
\label{tab:face-results}
\centering
\footnotesize \setlength{\tabcolsep}{1pt}
\begin{tabular}{c|cccccccc}
\hline
Method &MOTA &MOTP &MT[\%] &ML[\%] &FP &FN &IDS &FM \\
\hline
\hline
CEM                   &18.9 &71.4 &18.8 &37.4 &1185.3 &4095.3 &69.8 &100.3 \\
KSP                &32.8 &{\bf 74.0} &15.1 &32.2 &648.5  &3589.3 &70.0 &82.3 \\
DCT                  &37.6 &73.7 &25.5 &12.6 &1235.0 &2691.0 &66.3 &59.3  \\
DPMF                 &42.6 &73.7 &24.6 &14.3 &679.0  &2858.3 &62.8 &74.0  \\
FH\textsuperscript{2}T &45.8 &73.4 &27.4 &11.5 &742.3  &2634.0 &43.0 &57.3  \\
\hline
NT   &{\bf 53.1} &70.4 &{\bf 34.2} &{\bf 8.5}  &{\bf 648.5}  &{\bf 2292.8} &{\bf 37.5} &{\bf 36.3}    \\
\hline
\end{tabular}
\end{table}

As presented in Table \ref{tab:face-results}, we find that our approach achieves the best performance on almost all evaluation metrics except MOTP. Specifically, the NT method produces $7.3\%$ and $6.8\%$ larger average MOTA and MT scores, and $3.0\%$ lower average ML score, comparing to the second best FH\textsuperscript{2}T tracker. The evaluated sequences are recorded in the unconstrained scenes with fast motion, illumination variations, motion blurs and frequent occlusions. Since different degrees of dependencies among objects are considered, our method is able to exploit different types of motion patterns to improve the tracking performance, indicated by the consistent highest scores of almost all metrics (\ie, MOTA, MT, ML, FP, FN, IDS, and FM). Meanwhile, comparison with the state-of-the-art methods, our approach tracks the objects more robustly even when occlusions occur, indicated by the IDS, FM and FN scores. However, the linear interpolation is used in our method to estimate the occluded parts of the trajectories, which is not accurate enough to achieve good MOTP score, especially for crowded scenes containing non-linear motion patterns.

{\noindent \textbf{Running Time.}}
We implement the NT algorithm in C++ without any code optimization. To demonstrate the running time of NT, we run it five times using a single thread on a {\it laptop} with a $2.8$ GHz Intel processor and $16$ GB memory. Given the detections with the corresponding CNN features, the average speeds on the multi-pedestrian tracking dataset, MOT2016 dataset, and multi-face tracking dataset are $7.9$, $0.8$, and $9.0$ frame per second (FPS), respectively.

\section{Conclusions}
In this work, we propose a non-uniform hypergraph learning based near-online MOT method, which assembles different degrees of dependencies among tracklets in a unified objective. In contrast to previous graph or hypergraph based methods, our formulation exploit different high-degree cues among multiple tracklets in a computationally efficient way. Extensive experiments on several datasets, including the multi-pedestrian and multi-face tracking datasets, and MOT2016 benchmark, show that our method achieves comparable performance regarding to the state-of-the-arts. For future work, we plan to investigate and compare different optimization strategies to solve the dense structure searching problem on non-uniform hypergraphs.

\section{ Acknowledgments}
Dawei Du and Siwei Lyu are supported by US NSF IIS-1816227 and the National Natural Science Foundation of China under Grant 61771341.

\appendix

\section{The Proof of Calculating the Updating Step $\eta$.}
We present the proof of calculating the updating step $\eta$. As discussed in the paper, the objective of dense structure searching on non-uniform hypergraph is defined as
\begin{equation}
\begin{array}{ll}
\Theta({\bf y})=\sum_{d=1}^{{\it D}}\lambda_{d}\sum_{{\bf v}_{1:d}\in{\cal N}(v_s)} {\cal A}({\bf v}_{1:d})\prod_{j=1}^{d} y_{v_j}.
\end{array}
\end{equation}

We use the pairwise updating scheme to search the dense structures on the hypergraph to complete the tracking task. Specifically, we increase one component $y_p$ and decrease another one $y_q$ appropriately, to increase $\Theta({\bf y})$, \ie,
\begin{equation}
y_{l}^\prime = \left \{
\begin{array}{cl}
&y_l, \quad l \neq p, l\neq q; \\
&y_l+\eta, \quad l=p; \\
&y_l-\eta, \quad l=q,
\end{array}
\right.
\label{equ:update-strategy}
\end{equation}
where ${\bf y}^\prime=(y_1^\prime, \cdots, y_{|{\cal N}(v_s)|}^\prime)$ is the updated indicator variable in the optimization process, and $l=1,\cdots,|{\cal N}(v_s)|$.

The objective with the updated indicator variable is calculated as:
\begin{align}
&\Theta({\bf y}^\prime) \nonumber \\
&= \lambda_{1}\sum_{v_i\ne{p,q}} {\cal A}(v_i)y_{v_i} + \lambda_{1}{\cal A}(p)(y_{p}+\eta)
+ \lambda_{1}{\cal A}(q)(y_{q}-\eta) \nonumber \\
&+ \lambda_{2}\sum_{v_i,v_j\ne{p,q}} {\cal A}(v_i,v_j) y_{v_i}y_{v_j}
+ \lambda_{2}\sum_{v_i\ne{p,q}} {\cal A}(v_i, p) y_{v_i}(y_{p}+\eta) \nonumber \\
&+ \lambda_{2}\sum_{v_i\ne{p,q}} {\cal A}(v_i, q) y_{v_i}(y_{q}-\eta)
+ \lambda_{2}{\cal A}(p,q)(y_{p}+\eta)(y_{q}-\eta) \nonumber \\
&+ \sum_{d=3}^{{\it D}}\lambda_{d}\sum_{{\bf v}_{1:d}\ne{p,q}} {\cal A}({\bf v}_{1:d})\prod_{j=1}^{d} y_{v_j}\nonumber \\
&+ \sum_{d=3}^{{\it D}}\lambda_{d}\sum_{{\bf v}_{1:d-1}\ne{p,q}} {\cal A}({\bf v}_{1:d-1},p)(y_{p}+\eta)\prod_{j=1}^{d-1} y_{v_j} \nonumber \\
&+ \sum_{d=3}^{{\it D}}\lambda_{d}\sum_{{\bf v}_{1:d-1}\ne{p,q}} {\cal A}({\bf v}_{1:d-1},q)(y_{q}-\eta)\prod_{j=1}^{d-1} y_{v_j}  \nonumber \\
&+ \sum_{d=3}^{{\it D}}\lambda_{d}\sum_{{\bf v}_{1:d-2}\ne{p,q}} {\cal A}({\bf v}_{1:d-2},p,q)(y_{p}+\eta)(y_{q}-\eta)\prod_{j=1}^{d-2} y_{v_j}.
\end{align}

The difference of objective after updating is
\begin{align}
&\Delta\Theta({\bf y})=\Theta({\bf y}^\prime) - \Theta({\bf y}) = \Big(\lambda_{1}{\cal A}(p) - \lambda_{1}{\cal A}(q)\Big)\cdot\eta \nonumber \\
& + \Big(\lambda_{2}\sum_{v_i\ne{p}}{\cal A}(v_i, p) y_{v_i} - \lambda_{2}\sum_{v_i\ne{q}}{\cal A}(v_i, q) y_{v_i}\Big)\cdot\eta \nonumber \\
& - \lambda_{2}{\cal A}(p,q)\cdot\eta^2 + \Big(\sum_{d=3}^{{\it D}}\lambda_{d}\sum_{{\bf v}_{1:d-1}\ne{p}}{\cal A}({\bf v}_{1:d-1},p)\prod_{j=1}^{d-1}y_{v_j}\nonumber \\
& - \sum_{d=3}^{{\it D}}\lambda_{d}\sum_{{\bf v}_{1:d-1}\ne{q}}{\cal A}({\bf v}_{1:d-1},q)\prod_{j=1}^{d-1}y_{v_j} \Big)\cdot\eta \nonumber \\
&-\sum_{d=3}^{{\it D}}\lambda_{d}\sum_{{\bf v}_{1:d-2}\ne{p,q}}{\cal A}({\bf v}_{1:d-2},p,q)\prod_{j=1}^{d-2}y_{v_j}\cdot\eta^2 \nonumber \\
&= - \Big( \lambda_{2}{\cal A}(p,q) \nonumber \\ 
&+ \sum_{d=3}^{{\it D}}\lambda_{d}\sum_{{\bf v}_{1:d-2}\ne{p,q}}{\cal A}({\bf v}_{1:d-2},p,q)\prod_{j=1}^{d-2}y_{v_j} \Big) \cdot \eta^2 \nonumber \\
&+ \Big(\lambda_{1}{\cal A}(p) - \lambda_{1}{\cal A}(q) + \sum_{d=2}^{{\it D}}\lambda_{d}\sum_{{\bf v}_{1:d-1}\ne{p}}{\cal A}({\bf v}_{1:d-1},p)\prod_{j=1}^{d-1}y_{v_j}  \nonumber \\
&- \sum_{d=2}^{{\it D}}\lambda_{d}\sum_{{\bf v}_{1:d-1}\ne{q}}{\cal A}({\bf v}_{1:d-1},q)\prod_{j=1}^{d-1}y_{v_j} \Big) \cdot \eta.
\end{align}

Then, we rewrite the difference of objective as
\begin{equation}
\Delta\Theta({\bf y}) = \varphi_{p,q}({\bf y}) \cdot \eta^2 + (\phi_{p}({\bf y}) - \phi_{q}({\bf y}))\cdot \eta,
\end{equation}
where
\begin{align}
\varphi_{p,q}({\bf y}) &= -\lambda_2\cdot{\cal A}(p,q) \nonumber\\
                       &-\sum_{d=3}^{D}\lambda_{d}\sum_{{\bf v}_{1:d-2}\ne{p,q}} {\cal A}({\bf v}_{1:d-2},p,q)\prod_{j=1}^{d-2} y_{v_j}, \\
\phi_{p}({\bf y}) &= \lambda_{1}{\cal A}(p) \nonumber\\
&+\sum_{d=2}^{D}\lambda_{d}\sum_{{\bf v}_{1:d-1}\in{{\cal N}(v_s)}}{\cal A}({\bf v}_{1:d-1}, p)\prod_{j=1}^{d-1}y_{v_j}.
\end{align}

As discussed in the paper, we select an appropriate updating step $\eta$ to maximize the objective difference $\Delta\Theta({\bf y})$\footnote{When $\varphi_{p,q}({\bf y})={0}$ and $\phi_{p}({\bf y})=\phi_{q}({\bf y})$, we have $\Delta\Theta({\bf y})=0$. We can not select any $\eta$ to increase the objective. Thus, we ignore this case in discussion.}. Based on the updating strategy presented in \eqref{equ:update-strategy}, we have two constraints of $\eta$, \ie, $0\leq{y}^\prime_{p}=y_{p}+\eta\leq\frac{1}{\hat{\alpha}}$, and $0\leq{y}^\prime_{q}=y_{q}-\eta\leq\frac{1}{\hat{\alpha}}$. Since $0\leq{y}_{p}\leq\frac{1}{\hat{\alpha}}$ and $0\leq{y}_{q}\leq\frac{1}{\hat{\alpha}}$, we have $\eta\leq{y_{q}}$, and $\eta\leq\frac{1}{\hat{\alpha}}-y_{p}$. Notably, in general, we can assume $\phi_{p}({\bf y})\geq\phi_{q}({\bf y})$. When $\phi_{p}({\bf y})<\phi_{q}({\bf y})$, we can exchange indexes $p$ and $q$ to maximize $\Delta \Theta({\bf y})$.

In this way, we can select the updating step $\eta$ as follows:
\begin{itemize}
\item if $\varphi_{p,q}({\bf y})\geq{0}$, we have $\Delta\Theta({\bf y}) = \varphi_{p,q}({\bf y}) \cdot \eta^2 + (\phi_{p}({\bf y}) - \phi_{q}({\bf y}))\cdot \eta$. To maximize $\Delta\Theta({\bf y})$, we have to satisfy the constraints of $\eta$, \ie, $\eta\leq{y_{q}}$, and $\eta\leq\frac{1}{\hat{\alpha}}-y_{p}$. We set $\eta=\min(y_{q},\frac{1}{\hat{\alpha}}-y_{p})$.
\item if $\varphi_{p,q}({\bf y})<{0}$, we have $\Delta\Theta({\bf y})=\varphi_{p,q}({\bf y})\cdot\Big( \eta + \frac{\phi_{p}({\bf y}) - \phi_{q}({\bf y})}{2\cdot\varphi_{p,q}({\bf y})} \Big)^2 - \frac{\big(\phi_{p}({\bf y}) - \phi_{q}({\bf y})\big)^2}{4\cdot\varphi_{p,q}({\bf y})}$. To maximize $\Delta\Theta({\bf y})$ and satisfy the constraints of $\eta$, \ie, $\eta\leq{y_{q}}$, and $\eta\leq\frac{1}{\hat{\alpha}}-y_{p}$, we set $\eta=\min\Big(y_{q}, \frac{1}{\hat{\alpha}}-y_{p}, \frac{\phi_{q}({\bf y}) -\phi_{p}({\bf y})}{2\cdot\varphi_{p,q}({\bf y})}\Big)$.
\end{itemize}

\begin{figure*}
\centering
\includegraphics[width=1.0\linewidth]{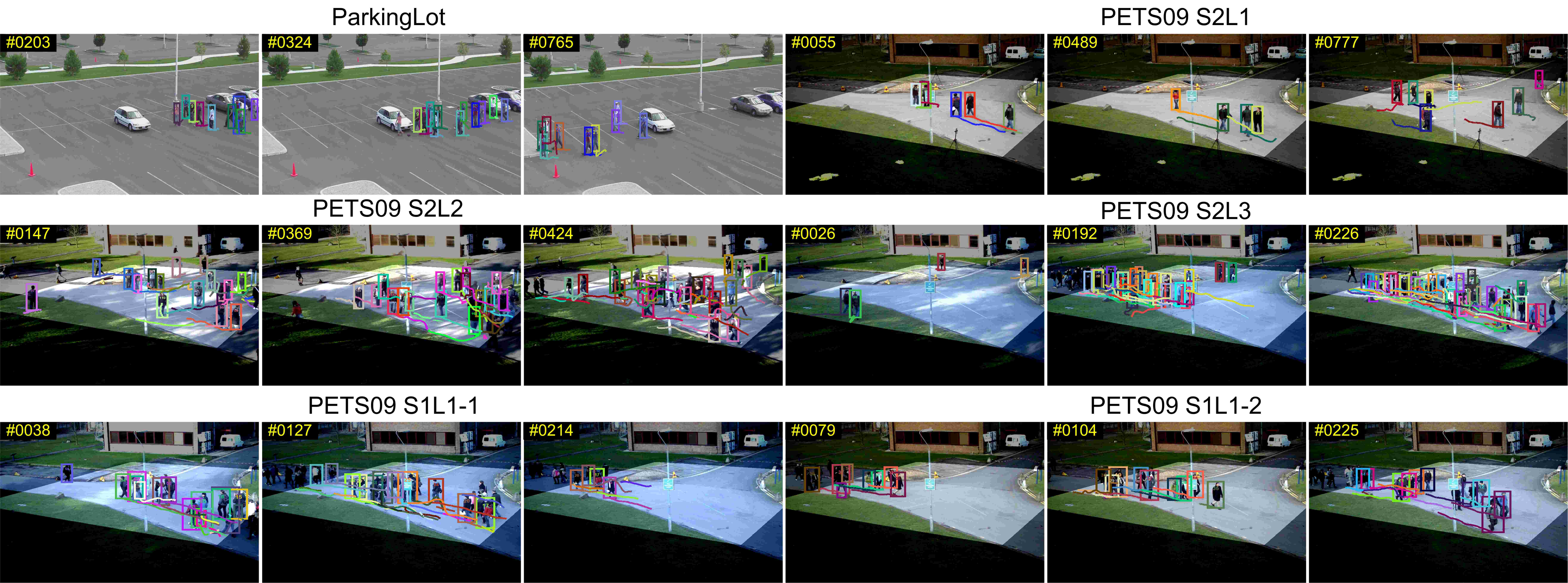}
\caption{Tracking results of the proposed NT tracker in multi-pedestrian tracking dataset.}
\label{fig:multi-pedestrian-results}
\end{figure*}

\section{Training Details of the Set to Set Recognition Model}
We fine-tune the GoogLeNet \cite{DBLP:conf/cvpr/SzegedyLJSRAEVR15} based set to set recognition model \cite{DBLP:conf/cvpr/LiuYO17} pre-trained on the ILSVRC CLS-LOC dataset \cite{DBLP:conf/nips/KrizhevskySH12} in the MOT16 training set to extract the CNN features of detections. Specifically, we divide the ground truth trajectories of pedestrians equally to form the two-view structure of \cite{DBLP:conf/cvpr/LiuYO17} in training. We optimize the network \cite{DBLP:conf/cvpr/LiuYO17} using the Stochastic Gradient Descent (SGD) algorithm with $0.9$ momentum and $0.0002$ weight decay on a Titan X GPU. We set the learning rate to $0.001$ for $120k$ iterations with a mini-batch of size $24$.

\section{Qualitative Tracking Results}
\begin{figure*}
\centering
\includegraphics[width=1.0\linewidth]{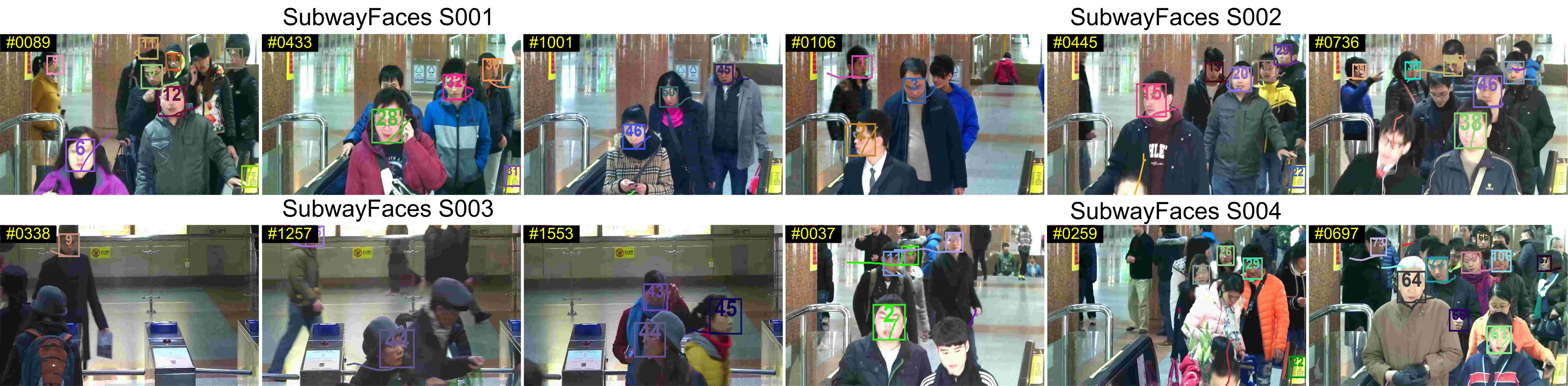}
\caption{Tracking results of the proposed NT tracker in multi-face tracking dataset.}
\label{fig:multi-face-results}
\end{figure*}

\begin{figure*}
\centering
\includegraphics[width=1.0\linewidth]{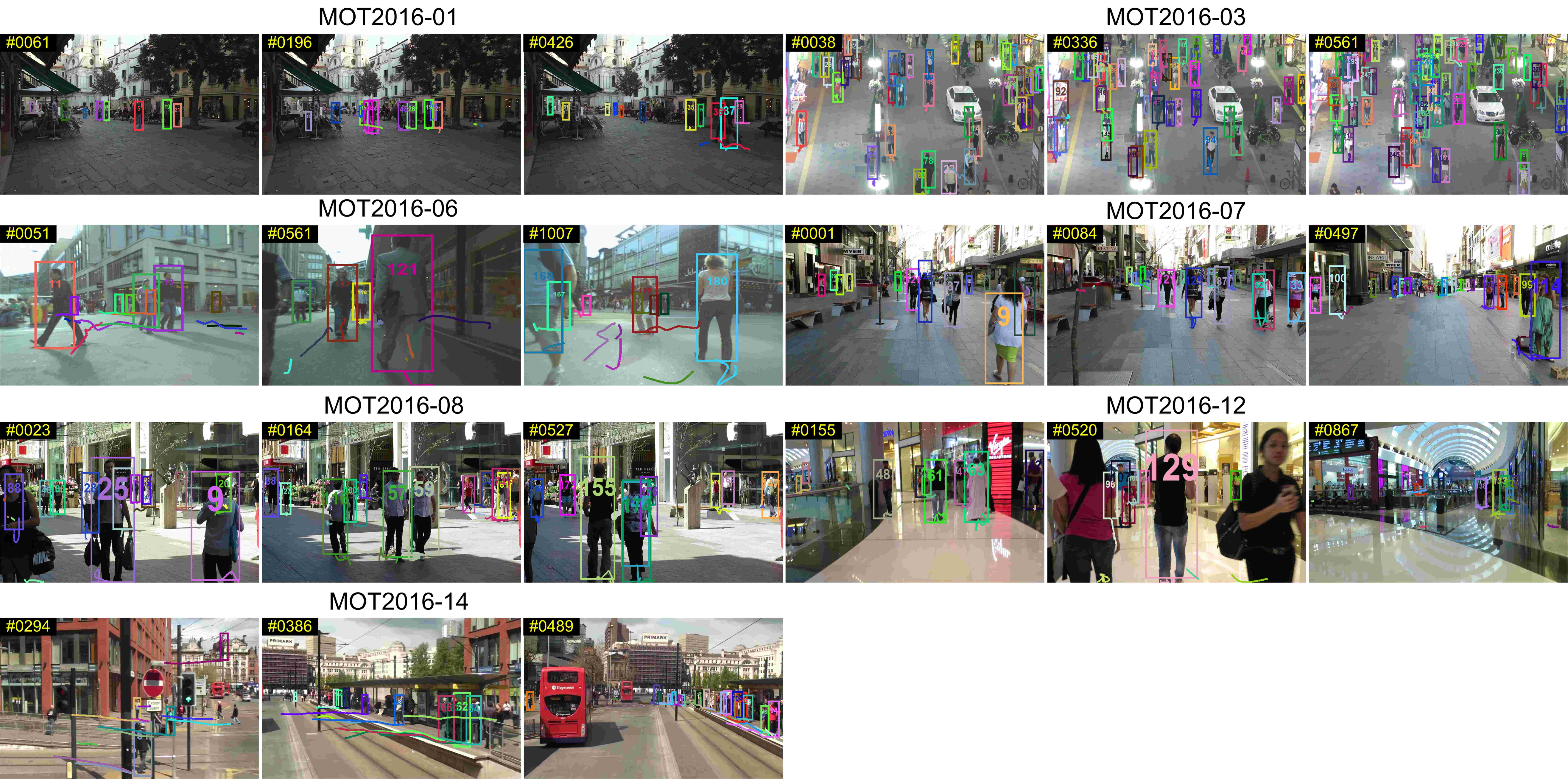}
\caption{Tracking results of the proposed NT tracker in the {\em test set} of MOT2016 benchmark.}
\label{fig:multi-mot2016-results}
\end{figure*}

We present some qualitative results of the proposed NT algorithm in Figure \ref{fig:multi-pedestrian-results}, Figure \ref{fig:multi-face-results}, and Figure \ref{fig:multi-mot2016-results}. More tracking results of our tracker are presented in the video demo. The proposed NT algorithm achieves good results mainly due to the introduction of non-uniform hypergraph learning in tracking task, which has much stronger descriptive power to accommodate different scenarios than the conventional graph or uniform hypergraph.

\small
\bibliography{references}
\bibliographystyle{aaai}

\end{document}